\documentclass[conference]{IEEEtran}
\IEEEoverridecommandlockouts
\usepackage{amsmath,amssymb,amsfonts}
\usepackage{algorithmic}
\usepackage{graphicx}
\usepackage{textcomp}
\usepackage{xcolor}
\def\BibTeX{{\rm B\kern-.05em{\sc i\kern-.025em b}\kern-.08em
    T\kern-.1667em\lower.7ex\hbox{E}\kern-.125emX}}

\usepackage{times}
\usepackage{epsfig}

\usepackage{color}
\usepackage{hyperref}
\hypersetup{
	colorlinks=true
}
\usepackage[figuresright]{rotating}
\usepackage{textcomp,subfigure,algorithm,multirow,upgreek,booktabs}
\usepackage{graphics,gensymb,setspace}

\usepackage{caption}
\usepackage{enumitem}
\setitemize{noitemsep,topsep=0pt,parsep=0pt,partopsep=0pt}
\setenumerate{noitemsep,topsep=0pt,parsep=0pt,partopsep=0pt}

\let\OLDthebibliography\thebibliography
\renewcommand\thebibliography[1]{
	\OLDthebibliography{#1}
	\setlength{\parskip}{0pt}
	\setlength{\itemsep}{0pt plus 0.4ex}
}

\begin{document}

\title{Attention-Guided Generative Adversarial Networks \\ for Unsupervised Image-to-Image Translation
}

\author{\IEEEauthorblockN{Hao Tang$^1$ \quad Dan Xu$^2$  \quad Nicu Sebe$^1$ \quad Yan Yan$^3$}
	\IEEEauthorblockA{$^1$Department of Information Engineering and Computer Science, University of Trento, Trento, Italy\\
		$^2$Department of Engineering Science, University of Oxford, Oxford, United Kingdom \\
		$^3$Department of Computer Science, Texas State University, San Marcos, USA\\
		{\tt\small \{hao.tang, niculae.sebe\}@unitn.it, danxu@robots.ox.ac.uk, y\_y34@txstate.edu
	}}}

\maketitle

\begin{abstract}
	
	The state-of-the-art approaches in Generative Adversarial Networks (GANs)  are able to learn a mapping function from one image domain to another with unpaired image data.
	However, these methods often produce artifacts and can only be able to convert low-level information, but fail to transfer high-level semantic part of images. 
	The reason is mainly that generators do not have the ability to detect the most discriminative semantic part of images, which thus makes the generated images with low-quality. 
	To handle the limitation, in this paper we propose a novel Attention-Guided Generative Adversarial Network (AGGAN), which can detect the most discriminative semantic object and minimize changes of unwanted part for semantic manipulation problems without using extra data and models.
	The attention-guided generators in AGGAN are able to produce attention masks via a built-in attention mechanism, and then fuse the input image with the attention mask to obtain a target image with high-quality.
	Moreover, we propose a novel attention-guided discriminator which only considers attended regions.
	The proposed AGGAN is trained by an end-to-end fashion with an adversarial loss, cycle-consistency loss, pixel loss and attention loss.
	Both qualitative and quantitative results demonstrate that our approach is effective to generate sharper and more accurate images than existing models. The code is available at~\url{https://github.com/Ha0Tang/AttentionGAN}.

\end{abstract}

\section{Introduction}

\begin{figure*}[!t] \tiny
	\centering
	\includegraphics[width=1\linewidth]{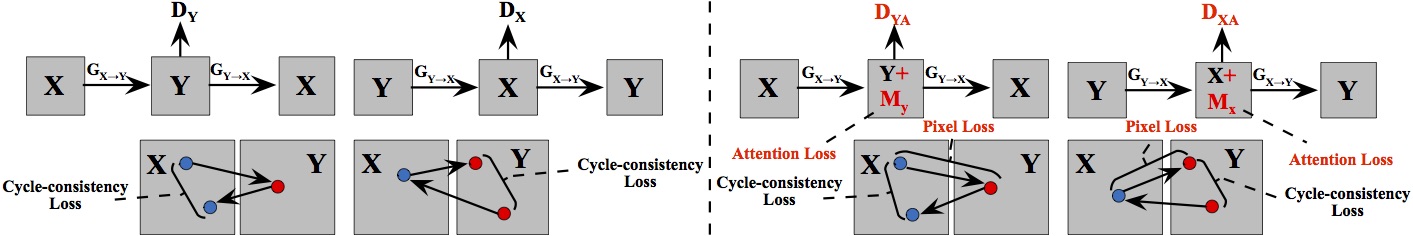}
	\caption{Comparison of previous frameworks, \emph{e.g.}, CycleGAN \cite{zhu2017unpaired}, DualGAN \cite{yi2017dualgan} and DiscoGAN \cite{kim2017learning} (Left), and the proposed AGGAN (Right). The contribution of AGGAN is that the proposed generators can produce the attention mask ($M_x$ and $M_y$) via the built-in attention module and then the produced attention mask and content mask mixed with the input image to obtain the targeted image. Moreover, we also propose two attention-guided discriminators $D_{XA}$, $D_{YA}$, which aim to consider only the attended regions. Finally, for better optimizing the proposed AGGAN, we employ pixel loss, cycle-consistency loss and attention loss.}
	\label{fig:task}
	\vspace{-0.6cm}
\end{figure*}

Recently, Generative Adversarial Networks (GANs)  \cite{goodfellow2014generative} have received considerable attention across many communities, \emph{e.g.}, computer vision, natural language processing,  audio and video processing.
GANs are generative models, which are particularly designed for image generation task.
Recent works in computer vision, image processing and computer graphics have produced powerful translation systems in supervised settings such as Pix2pix \cite{isola2016image}, where the image pairs are required. 
However, the paired training data are usually difficult and expensive to obtain. 
Especially, the input-output pairs for images tasks such as artistic stylization can be even more difficult to acquire since the desired output is quite complex, typically requiring artistic authoring. 
To tackle this problem, CycleGAN~\cite{zhu2017unpaired}, DualGAN~\cite{yi2017dualgan} and DiscoGAN~\cite{kim2017learning} provide an insight, in which the models can learn the mapping from one image domain to another one with unpaired image data.

Despite these efforts, image-to-image translation, \emph{e.g.}, converting a neutral expression to a happy expression, remains a challenging problem due to the fact that the facial expression changes are non-linear, unaligned and vary conditioned on the appearance of the face. 
Moreover, most previous models change unwanted objects during the translation stage and can also be easily affected by background changes. 
In order to address these limitations, 
Liang \textit{et al.} propose the ContrastGAN~\cite{liang2017generative}, which uses object mask annotations from each dataset.
In ContrastGAN, it first crops a part in the image according to the masks, and then makes translations and finally pastes it back. 
Promising results have been obtained from it, however it is hard to collect training data with object masks.
More importantly, we have to make an assumption that the object shape should not change after applying semantic modification. 
Another option is to train an extra model to detect the object masks and fit them into the generated image patches \cite{chen2018attention,kastaniotis2018attention}. 
In this case, we need to increase the number of parameters of our network, which consequently increases the training complexity both in time and space.

To overcome the aforementioned issues, in this paper we propose a novel Attention-Guided Generative Adversarial Network (AGGAN) for the image translation problem without using extra data and models.
The proposed AGGAN comprises of two generators and two discriminators, which is similar with CycleGAN \cite{zhu2017unpaired}. 
Fig.~\ref{fig:task} illustrates the differences between previous representative works and the proposed AGGAN. 
Two attention-guided generators in the proposed AGGAN have built-in attention modules, which can disentangle the discriminative semantic object and the unwanted part by producing a attention mask and a content mask.
Then we fuse the input image with new patches produced through the attention mask to obtain high-quality results. 
We also constrain generators with pixel-wise and cycle-consistency loss function, which forces the generators to reduce changes. 
Moreover, we propose two novel attention-guided discriminators which aims to consider only the attended regions.
The proposed AGGAN is trained by an end-to-end fashion, and can produce attention mask, content mask and targeted images at the same time.
Experimental results on four public available datasets demonstrate that the proposed AGGAN is able to produce higher-quality images compared with the state-of-the-art methods. 

The contributions of this paper are summarized as follows:
\begin{itemize} [leftmargin=*]
	\item We propose a novel Attention-Guided Generative Adversarial Network (AGGAN) for unsupervised image-to-image translation.
	\item We propose a novel generator architecture with built-in attention mechanism, which can detect the most discriminative semantic part of images in different domains.
	\item We propose a novel attention-guided discriminator which only consider the attended regions. Moreover, the proposed attention-guided generator and discriminator can be easily used to other GAN models.
	\item Extensive results demonstrate that the proposed AGGAN can generate sharper faces with clearer details and more realistic expressions compared with baseline models. 
\end{itemize}

\section{Related Work}

\noindent \textbf{Generative Adversarial Networks (GANs)} \cite{goodfellow2014generative} are powerful generative models, which have achieved impressive results on different computer vision tasks, \emph{e.g.},
image generation \cite{brock2018large,park2017transformation,huang2017beyond}, image editing \cite{shen2016learning,shu2017neural} and image inpainting
\cite{li2017generative,iizuka2017globally}.
In order to generate meaningful images that meet user requirement, Conditional GAN (CGAN)~\cite{mirza2014conditional} is proposed where the conditioned information is employed to guide the image generation process.
The conditioned information can be discrete labels \cite{perarnau2016invertible}, text \cite{mansimov2015generating,reed2016generative}, object keypoints \cite{reed2016learning}, human skeleton \cite{tang2018gesturegan} and reference images \cite{isola2016image}. 
CGANs using a reference images as conditional information have tackled a lot of problems, \emph{e.g.}, text-to-image translation~\cite{mansimov2015generating}, image-to-image translation \cite{isola2016image} and video-to-video translation \cite{wang2018video}.

\begin{figure*}[!t] \tiny
	\centering
	\includegraphics[width=0.85\linewidth]{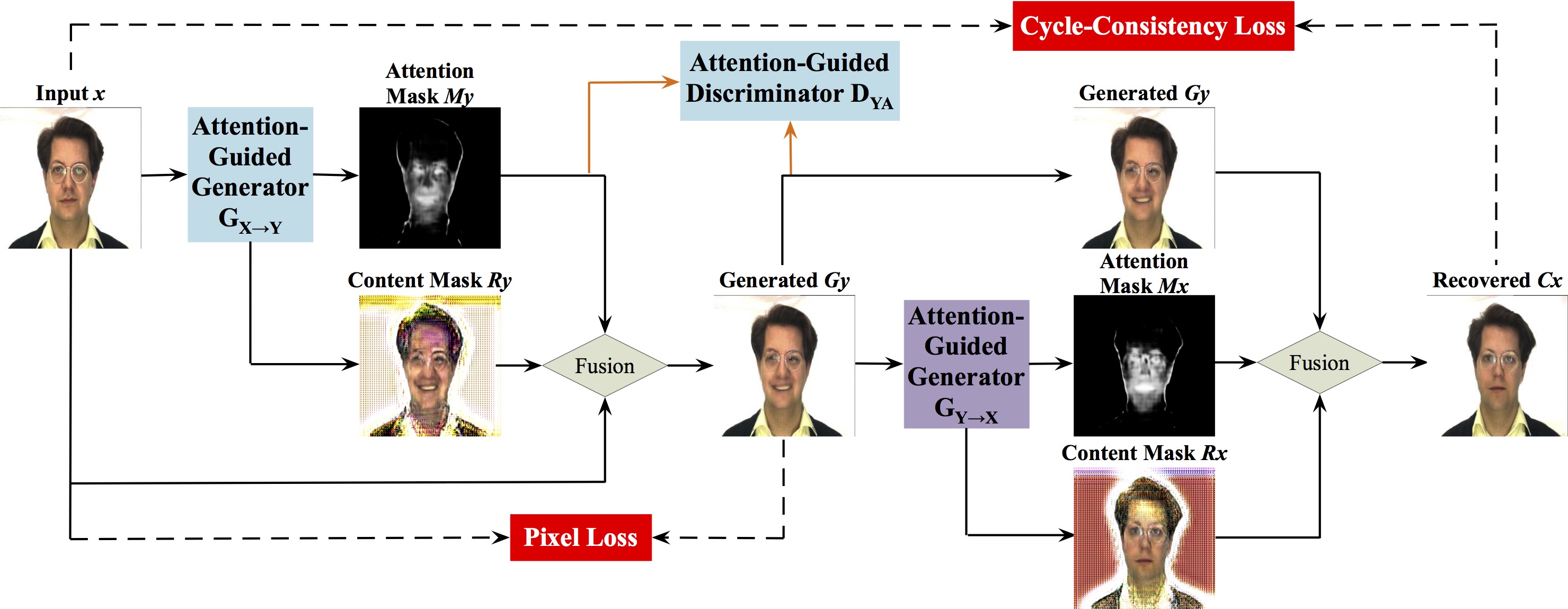}
	\caption{The framework of the proposed AGGAN. Because of the space limitation, we only show one mapping in this figure, \emph{i.e.}, $x {\rightarrow} [M_y, R_y, G_y] {\rightarrow} C_x {\approx} x$. We also have the other mapping, \emph{i.e.}, $y {\rightarrow} [M_x, R_x, G_x] {\rightarrow} C_y {\approx} y$. The attention-guided generators have built-in attention mechanism, which can detect the most discriminative part of images. After that we mix the input image, content mask and the attention mask to synthesize the targeted image. Moreover, to distinguish only the most discriminative content, we also propose a attention-guided discriminator $D_{YA}$. Note that our systems does not require supervision, \emph{i.e.}, no pairs of images of the same person with different expressions.}
	\label{fig:excyclegan}
	\vspace{-0.6cm}
\end{figure*}

\noindent \textbf{Image-to-Image Translation} models learns a translation function using CNNs. 
Pix2pix \cite{isola2016image} is a conditional framework using a CGAN to learn a mapping function from input to output images. 
Similar ideas have also been applied to many other tasks, such as generating photographs from sketches \cite{sangkloy2016scribbler} or vice versa \cite{tang2019attribute}. 
However, most of the tasks in the real world suffer from the constraint of having few or none of the paired input-output samples available.
To overcome this limitation, unpaired image-to-image translation task has been proposed.
Different from the prior works, unpaired image translation task try to learn the mapping function without the requirement of paired training data.
Specifically, CycleGAN \cite{zhu2017unpaired} learns the mappings between two image domains (\emph{i.e.}, a source domain $X$ to a target domain $Y$) instead of the paired images.
Apart from CycleGAN, many other GAN variants are proposed to tackle the cross-domain problem.
For example, to learn a common representation across domains, CoupledGAN \cite{liu2016coupled} uses a weight-sharing strategy. 
The work of \cite{taigman2016unsupervised} utilizes some certain shared content features between input and output even though they may differ in style. 
Kim \textit{et al.} \cite{kim2017learning} propose a method based on GANs that learns to discover relations between different domains.
A model which can learn object transfiguration from two unpaired sets of images is presented in \cite{zhou2017genegan}.
Tang \textit{et al.} \cite{tang2018dual} propose G$^2$GAN, which is a robust and scalable approach allowing to perform unpaired image-to-image translation for multiple domains. 
However,  those models can be easily affected by unwanted content and cannot focus on the most discriminative semantic part of images during translation stage.

\noindent \textbf{Attention-Guided Image-to-Image Translation.}
In order to fix the aforementioned limitations, Liang \textit{et al.} propose ContrastGAN \cite{liang2017generative}, which uses the object mask annotations from each dataset as extra input data.
In this method, we have to make an assumption that after applying semantic changes an object shape does not change.
Another method is to train another segmentation or attention model and fit it to the system.
For instance, 
Mejjati \textit{et al.} \cite{mejjati2018unsupervised} propose an attention mechanisms that are jointly trained with the generators and discriminators.
Chen \textit{et al.} propose AttentionGAN \cite{chen2018attention}, which uses an extra attention network to generate attention maps, so that major attention can be paid to objects of interests.
Kastaniotis \textit{et al.} \cite{kastaniotis2018attention} present ATAGAN, which use a teacher network to produce attention maps.
Zhang \textit{et al.} \cite{zhang2018self} propose the Self-Attention Generative Adversarial Networks (SAGAN) for image generation task.
Qian \textit{et al.} \cite{qian2018attentive} employ a recurrent network to generate visual attention first and then transform a raindrop degraded image into a clean one.
Tang \textit{et al.}~\cite{tang2019selection} propose a novel Multi-Channel Attention Selection GAN for the challenging cross-view image translation task.
Sun \textit{et al.}~\cite{sun2018mask} generate a facial mask by using FCN \cite{long2015fully} for face attribute manipulation.

All these aforementioned methods employ extra networks or data to obtain attention masks, which increases the number of parameters, training time and storage space of the whole system.
In this work, we propose the Attention-Guided Generative Adversarial Network (AGGAN), which can produce attention masks by the generators. 
For this purpose, we embed an attention method to the vanilla generator which means that we do not need any extra models to obtain the attention masks of objects of interests.

\section{Method}

We first start with the attention-guided generator and discriminator of the proposed Attention-Guided Generative Adversarial Network (AGGAN), and then introduce the loss function for better optimization of the model.
Finally we present the implementation details of the whole model including network architecture and training procedure.

\subsection{Attention-Guided Generator}
GANs~\cite{goodfellow2014generative} are composed of two competing modules, \emph{i.e.}, the generator $G_{X\rightarrow Y}$ and the discriminator $D_Y$ (where $X$ and $Y$ denote two different image domains), which are iteratively trained competing against with each other in the manner of two-player minimax.
More formally, let $x_i{\in} X$ and $y_j{\in} Y$ denote the training images in source and target image domain, respectively (for simplicity, we usually omit the subscript $i$ and~$j$).
For most current image translation models, \emph{e.g.}, CycleGAN \cite{zhu2017unpaired} and DualGAN \cite{yi2017dualgan}, they include two mappings $G_{X\rightarrow Y}{:} x {\rightarrow} G_y$ and $G_{Y\rightarrow X}{:} y {\rightarrow} G_x$, and two corresponding adversarial discriminators $D_X$ and $D_Y$.
The generator $G_{X\rightarrow Y}$ maps $x$ from the source domain to the generated image $G_y$ in the target domain $Y$ and tries to fool the discriminator $D_Y$, whilst the $D_Y$ focuses on improving itself in order to be able to tell whether a sample is a generated sample or a real data sample.
Similar to~$G_{Y\rightarrow X}$ and ~$D_X$.

While for the proposed AGGAN, we intend to learn two mappings between domains $X$ and $Y$ via two generators with built-in attention mechanism, \emph{i.e.}, $G_{X\rightarrow Y}{:} x {\rightarrow} [M_y, R_y, G_y]$ and $G_{Y\rightarrow X}{:} y {\rightarrow} [M_x, R_x, G_x]$, where $M_x$ and $M_y$ are the attention masks of images $x$ and $y$, respectively; 
$R_x$ and $R_y$ are the content masks of images $x$ and $y$, respectively;
$G_x$ and $G_y$ are the generated images.
The attention masks $M_x$ and $M_y$ define a per pixel intensity specifying to which extend each pixel of the content masks $R_x$ and $R_y$ will contribute in the final rendered image.
In this way, the generator does not need to render static elements, and can focus exclusively on the pixels defining the facial movements, leading to sharper and more realistic synthetic images.
After that, we fuse input image $x$ and the generated attention mask $M_y$, and the content mask $R_y$ to obtain the targeted image $G_y$. 
Through this way, we can disentangle the most discriminative semantic object and unwanted part of images.
In Fig.~\ref{fig:excyclegan}, the attention-guided generators focus only on those regions of the image that are responsible of generating the novel expression such as eyes and mouth, and keep the rest parts of the image such as hair, glasses, clothes untouched.
The higher intensity in the attention mask means the larger contribution for changing the expression.

To focus on the discriminative semantic parts in two different domains, we specifically designed two generators with built-in attention mechanism.
By using this mechanism, generators can generate attention masks in two different domains. 
The input of each generator is a three-channel image, and the outputs of each generator are a attention mask and a content mask.
Specifically, the input image of $G_{X\rightarrow Y}$ is $x{\in} \mathbb{R}^{H\times W \times 3}$, and the outputs are the attention mask $M_y {\in} \{0,...,1\}^{H\times W}$ and content mask $R_y {\in} \mathbb{R}^{H\times W \times 3}$.
Thus, we use the following formulation to calculate the final image~$G_y$,
\begin{equation}
\begin{aligned}
G_y = R_y * M_y + x * (1 - M_y),
\end{aligned}
\label{eqn:atten}
\end{equation}  
where attention mask $M_y$ is copied into three-channel for multiplication purpose.
The formulation for generator $G_{Y\rightarrow X}$ and input image $y$ is $G_x {=} R_x * M_x {+} y * (1 {-} M_x)$.
Intuitively, attention mask $M_y$ enables some specific areas where facial muscle changed to get more focus, applying it to the content mask $R_y$ can generate images with clear dynamic area and unclear static area.
After that, what left is to enhance the static area, which should be similar between the generated image and the original real image. 
Therefore we can enhance the static area (basically it refers to background area) in the original real image $(1 {-} M_y )*x$ and merge it to $R_y {*} M_y$ to obtain final result $R_y {*} M_y + x {*} (1 {-} M_y)$.

\subsection{Attention-Guided Discriminator}
Eq.~\eqref{eqn:atten} constrains the generators to act only on the attended regions. 
However, the discriminators currently consider the whole image.
More specifically, the vanilla discriminator $D_Y$ takes the generated image $G_y$ or the real image $y$ as input and tries to distinguish them.
Similar to discriminator $D_X$, which tries to distinguish the generated image $G_x$ and the real image $x$.  
To add attention mechanism to the discriminator so that it only considers attended regions.
We propose two attention-guided discriminators.
The attention-guided discriminator is structurally the same with the vanilla discriminator but it also takes the attention mask as input.
For attention-guided discriminator $D_{YA}$, which tries to distinguish the fake image pairs $[M_y, G_y]$ and the real image pairs $[M_y, y]$.
Similar to $D_{XA}$, which tries to distinguish the fake image pairs $[M_x, G_x]$ and the real image pairs $[M_x, x]$.
In this way, discriminators can focus on the most discriminative content.

\subsection{Optimization Objective}

\noindent \textbf{Vanilla Adversarial Loss} $\mathcal{L}_{GAN}(G_{X\rightarrow Y}, D_Y)$ \cite{goodfellow2014generative} can be formulated as follows:
\begin{equation}
\begin{aligned}
& \mathcal{L}_{GAN}(G_{X\rightarrow Y}, D_Y) =  \mathbb{E}_{y\sim{p_{\rm data}}(y)}\left[ \log D_Y(y)\right] + \\
&  \mathbb{E}_{x\sim{p_{\rm data}}(x)}[\log (1 - D_Y(G_{X\rightarrow Y}(x)))].
\end{aligned}
\label{equ:basicgan}
\end{equation}
$G_{X\rightarrow Y}$ tries to minimize the adversarial loss objective $\mathcal{L}_{GAN}(G_{X\rightarrow Y}, D_Y)$ while $D_Y$ tries to maximize it.
The target of $G_{X\rightarrow Y}$ is to generate an image $G_y{=}G_{X\rightarrow Y}(x)$ that looks similar to the images from domain $Y$, while $D_Y$ aims to distinguish between the generated images $G_{X\rightarrow Y}(x)$ and the real images $y$.
A similar adversarial loss of Eq.~ \eqref{equ:basicgan} for mapping function $G_{Y\rightarrow X}$ and its discriminator $D_X$ is defined as 
$\mathcal{L}_{GAN}(G_{Y\rightarrow X}, D_X)
=  \mathbb{E}_{x\sim{p_{\rm data}}(x)}[\log D_X(x)] 
+ \mathbb{E}_{y\sim{p_{\rm data}}(y)}[\log (1 - D_X(G_{Y\rightarrow X}(y)))].$

\noindent \textbf{Attention-Guided Adversarial Loss.}
We propose the attention-guided adversarial loss for training the attention-guide discriminators.
The min-max game between the attention-guided discriminator $D_{YA}$ and the generator $G_{X\rightarrow Y}$ is performed through the following objective functions:
\begin{equation}
\begin{aligned}
& \mathcal{L}_{AGAN}(G_{X\rightarrow Y}, D_{YA}) =  \mathbb{E}_{y\sim{p_{\rm data}}(y)}\left[ \log D_{YA}([M_y, y])\right] + \\
&  \mathbb{E}_{x\sim{p_{\rm data}}(x)}[\log (1 - D_{YA}([M_y, G_{X\rightarrow Y}(x)]))],
\end{aligned}
\label{equ:agan}
\end{equation}
where $D_{YA}$ aims to distinguish between the generated image pairs $[M_y, G_{X\rightarrow Y}(x)]$ and the real image pairs $[M_y,y]$.
We also have another  loss $\mathcal{L}_{AGAN}(G_{Y\rightarrow X}, D_{XA})$ for discriminator $D_{XA}$ and the generator $G_{Y\rightarrow X}$.

\noindent \textbf{Cycle-Consistency Loss.}
Note that  CycleGAN \cite{zhu2017unpaired} and DualGAN \cite{yi2017dualgan} are different from Pix2pix \cite{isola2016image} as the training data in those models are unpaired. 
The cycle-consistency loss can be used to enforce forward and backward consistency.
The cycle-consistency loss can be regarded as ``pseudo'' pairs of training data 
even though we do not have the corresponding data in the target domain which corresponds to the input data from the source domain.
Thus, the loss function of cycle-consistency can be defined as:
\begin{equation}
\begin{aligned}
&  \mathcal{L}_{cycle}(G_{X\rightarrow Y}, G_{Y\rightarrow X}) = \\
& \mathbb{E}_{x\sim{p_{\rm data}}(x)}[\Arrowvert G_{Y\rightarrow X}(G_{X\rightarrow Y}(x))-x\Arrowvert_1]  + \\
& \mathbb{E}_{y\sim{p_{\rm data}}(y)}[\Arrowvert G_{X\rightarrow Y}(G_{Y\rightarrow X}(y))-y\Arrowvert_1].
\end{aligned}
\label{equ:cycleganloss}
\end{equation}
The reconstructed images $C_x{=} G_{Y\rightarrow X}(G_{X\rightarrow Y}(x))$ are closely matched to the input image $x$, and similar to $G_{X\rightarrow Y}(G_{Y\rightarrow X}(y))$ and image $y$.

\noindent \textbf{Pixel Loss.}
To reduce changes and constrain generators, we adopt pixel loss between the input images and the generated images.
We express this loss as:
\begin{equation}
\begin{aligned}
\mathcal{L}_{pixel}(G_{X\rightarrow Y}, G_{Y\rightarrow X}) = & \mathbb{E}_{x\sim{p_{\rm data}}(x)}[\Arrowvert G_{X\rightarrow Y}(x)-x\Arrowvert_1]  + \\
& \mathbb{E}_{y\sim{p_{\rm data}}(y)}[\Arrowvert G_{Y\rightarrow X}(y)-y\Arrowvert_1].
\end{aligned}
\label{equ:pixelloss}
\end{equation}
We adopt $L1$ distance as loss measurement in pixel loss.
Note that the pixel loss usually used in the paired image-to-image translation models such as Pix2pix \cite{isola2016image}.
While we use it in our AGGAN for unpaired image-to-image translation task.

\noindent \textbf{Attention Loss.}
When training our AGGAN we do not have ground-truth annotation for the attention masks.
They are learned from the resulting gradients of the attention-guided discriminators and the rest of the losses. 
However, the attention masks can easily saturate to 1 which makes the attention-guided generator has no effect as indicated in~GANimation~\cite{pumarola2018ganimation}. 
To prevent this situation, we perform a Total Variation Regularization over attention masks $M_y$ and $M_x$. 
The attention loss of mask $M_x$ therefore can be defined as:
\begin{equation}
\begin{aligned}
\mathcal{L}_{tv}(M_x) = & \sum_{w,h=1}^{W,H} \left| M_x(w+1, h, c) - M_x(w, h, c)  \right| +  \\ 
&\left| M_x(w, h+1, c) - M_x(w, h, c) \right|,
\end{aligned}
\label{eqn:tv}
\end{equation}
where $W$ and $H$ are the width and height of $M_x$.

\noindent \textbf{Full Objective.}
Thus, the complete objective loss of AGGAN can be formulated as follows:
\begin{equation}
\begin{aligned}
& \mathcal{L}(G_{X \rightarrow Y}, G_{Y \rightarrow X}, D_X, D_Y, D_{XA}, D_{YA}) = \\
& \lambda_{gan}*[\mathcal{L}_{GAN}(G_{X\rightarrow Y}, D_Y) + \mathcal{L}_{GAN}(G_{Y\rightarrow X}, D_X) + \\
& \mathcal{L}_{AGAN}(G_{X\rightarrow Y}, D_{YA}) + \mathcal{L}_{AGAN}(G_{Y\rightarrow X}, D_{XA})] + \\ 
& \lambda_{cycle} * \mathcal{L}_{cycle}(G_{X \rightarrow Y}, G_{Y \rightarrow X}) + \\
& \lambda_{pixel} * \mathcal{L}_{pixel}(G_{X\rightarrow Y}, G_{Y\rightarrow X}) {+} \lambda_{tv} * [\mathcal{L}_{tv}(M_x) {+} \mathcal{L}_{tv}(M_y)].
\end{aligned}
\label{eqn:allloss}
\end{equation}
where $\lambda_{gan}$, $\lambda_{cycle}$, $\lambda_{pixel}$ and $\lambda_{tv}$ are parameters controlling the relative relation of objectives terms.
We aim to solve:
\vspace{-.2cm}
\begin{equation}\small
\begin{aligned}
& G_{X\rightarrow Y}^\ast, G_{Y \rightarrow X}^\ast = \\
& \arg \mathop{\min}\limits_{\substack{G_{Y\rightarrow X}, \\G_{X\rightarrow Y}}} \mathop{\max}\limits_{\substack{D_X, D_Y, \\D_{XA}, D_{YA}}} \mathcal{L}(G_{X \rightarrow Y}, G_{Y \rightarrow X}, D_X, D_Y, D_{XA}, D_{YA}).
\end{aligned}
\vspace{-.2cm}
\end{equation}

\subsection{Implementation Details}
\noindent \textbf{Network Architecture.} For fair comparison, we use the generator architecture from CycleGAN \cite{zhu2017unpaired}.
We have slightly modified it for our task and the network architecture of the proposed generators is, 
$[c7s1\_64, d128, d256, R256, R256, R256, R256, R256, R256, $\\ $u128, u64, c7s1\_4]$,
where $c7s1\_k$ denotes a $7 {\times} 7$ Convolution-BatchNorm-ReLU layer with $k$ filters and stride 1. 
$dk$ denotes a $3 {\times} 3$ Convolution-BatchNorm-ReLU layer with $k$ filters and stride 2. 
$Rk$ represents a residual block that contains two $3 {\times} 3$ convolutional layers with stride 1 and the same number of filters on both
layer. 
$uk$ denotes a $3 {\times} 3$ fractional-strided-Convolution-BatchNorm-ReLU layer with $k$ filters and stride $1/2$.
The generator takes an 3-channel RGB image as input and outputs a single-channel attention mask and a 3-channel content mask. 

For the vanilla discriminator, we employ discriminator architecture in \cite{isola2016image,zhu2017unpaired}, which is denoted as $[C64, C128, C256, C512, C512]$, where $Ck$ denotes a $4 {\times} 4$ Convolution-BatchNorm-LReLU layer with $k$ filters and stride 2. 
The differences between \cite{isola2016image,zhu2017unpaired} are that the BatchNorm is used for the first $C64$.
And for the last $C512$, the stride is change to 1 and BatchNorm is not adopted.
After the end of the discriminator architecture, an adaptive average pooling layer and a convolution layer are applied to produce the final 1 dimensional output. 
For comparing the vanilla discriminator and the proposed attention-guided discriminator, we employ the same architecture as the proposed attention-guided discriminator except the attention-guided discriminator takes a attention mask and an image as inputs while the vanilla discriminator only takes an image as input.

\noindent \textbf{Training Strategy.} 
We follow the standard optimization method from  \cite{goodfellow2014generative} to optimize the proposed AGGAN, \emph{i.e.}, we alternate between one gradient descent
step on generators, then one step on discriminators.
The proposed AGGAN is trained end-to-end fashion.
Moreover, in order to slow down the rate of discriminators relative to generators we divide the objective by 2 while optimizing discriminators.
We use a least square loss \cite{mao2016multi} to stabilize our model during training procedure similar to CycleGAN.
The least square loss is more stable than the negative log likelihood objective in Eq.~\eqref{equ:basicgan} and more faster than Wasserstein GAN (WGAN) \cite{arjovsky2017wasserstein} to converge.
We also use a history of generated images to update discriminators similar to CycleGAN.
Finally, we adopt a curriculum strategy for training stage, which makes we have a strong GAN loss at the beginning of training time. 
Eq.~\eqref{eqn:allloss} then becomes,
\begin{equation}
\begin{aligned}
& \mathcal{L}(G_{X \rightarrow Y}, G_{Y \rightarrow X}, D_X, D_Y, D_{XA}, D_{YA}) = \\
& [\lambda_{cycle} * \mathcal{L}_{cycle}(G_{X \rightarrow Y}, G_{Y \rightarrow X}) + \\
& \lambda_{pixel} * \mathcal{L}_{pixel}(G_{X\rightarrow Y}, G_{Y\rightarrow X})] * r + \\
& \{\lambda_{gan}*[\mathcal{L}_{GAN}(G_{X\rightarrow Y}, D_Y) + \mathcal{L}_{GAN}(G_{Y\rightarrow X}, D_X) + \\
& \mathcal{L}_{AGAN}(G_{X\rightarrow Y}, D_{YA}) + \mathcal{L}_{AGAN}(G_{Y\rightarrow X}, D_{XA})] + \\
& \lambda_{tv} * [\mathcal{L}_{tv}(M_x) + \mathcal{L}_{tv}(M_y)]\} * (1-r),  
\end{aligned}
\label{eqn:curriculum}
\end{equation}
where $r$ is a curriculum parameter to control the relation between GAN loss and reconstruction loss (\emph{i.e}, cycle-consistency loss and pixel loss) during curriculum period.

\section{Experimental Results}
\begin{figure*}[!t] \tiny
	\centering
	\includegraphics[width=0.97\linewidth]{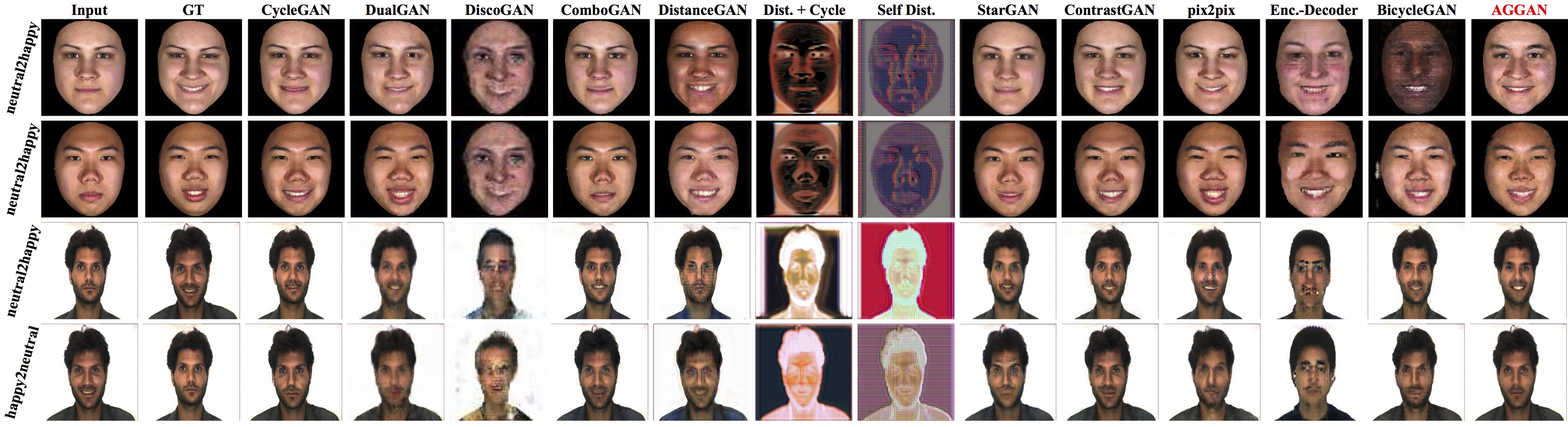}
	\caption{Comparison with different baselines on  Bu3dfe (Top) and AR Face (Bottom) datasets.}
	\label{fig:bu3dfe_ar}
	\vspace{-0.4cm}
\end{figure*}

\begin{figure*}[!t] \tiny
	\centering
	\includegraphics[width=0.97\linewidth]{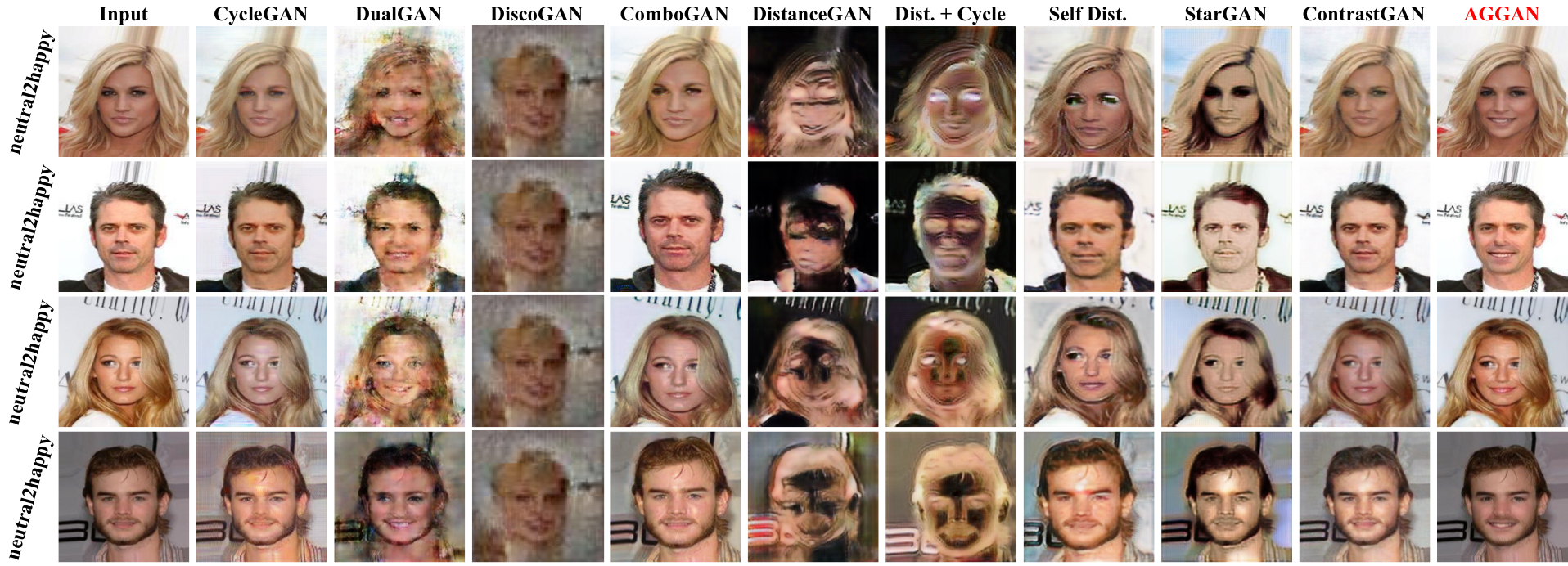}
	\caption{Comparison with different baselines on CelebA dataset.}
	\label{fig:celeba}
	\vspace{-0.4cm}
\end{figure*}

\begin{figure}[!t] \tiny
	\centering
	\includegraphics[width=1\linewidth]{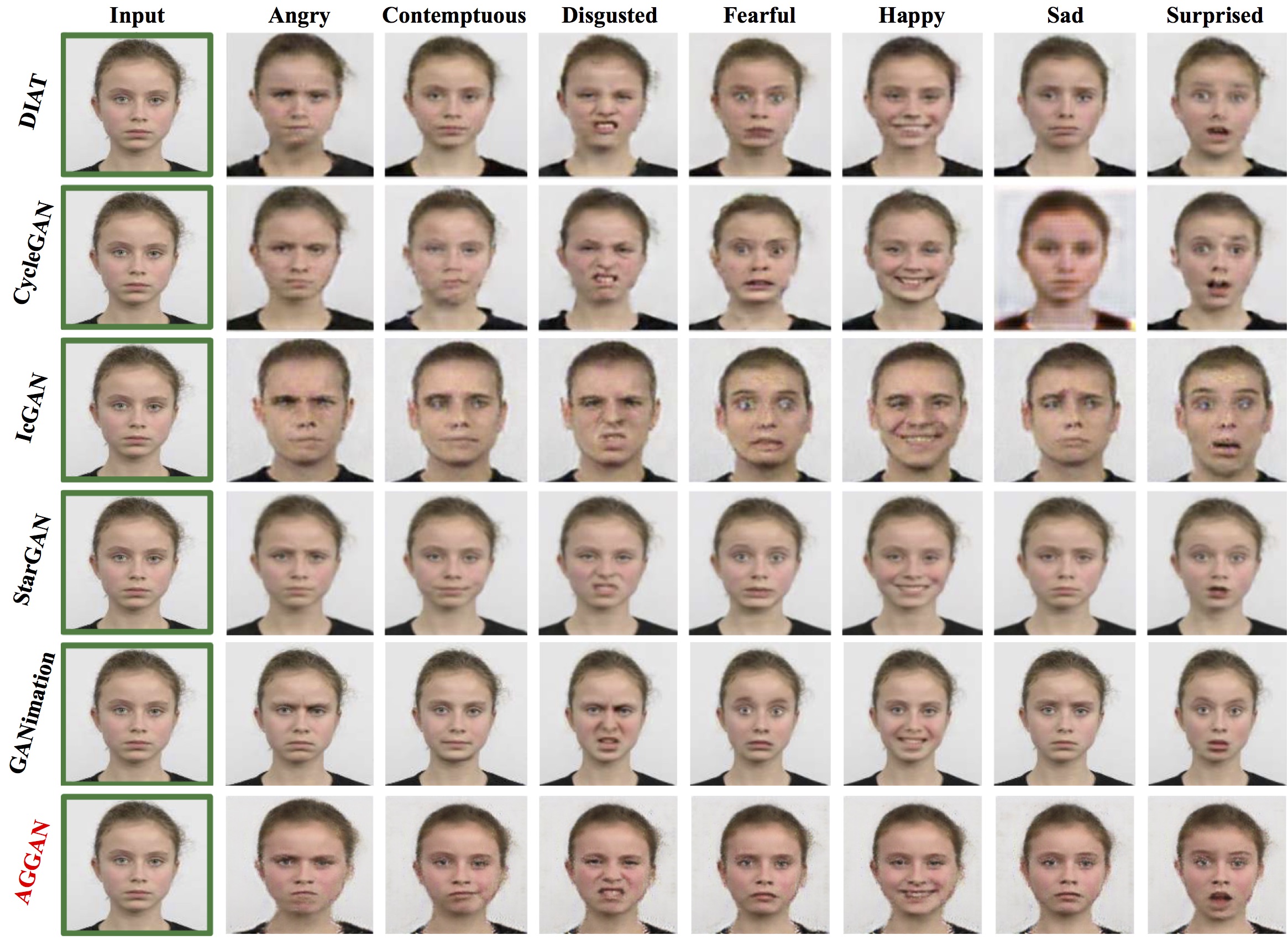}
	\caption{Comparison with baselines on RaFD dataset.}
	\label{fig:rafd}
	\vspace{-0.6cm}
\end{figure}

\subsection{Experimental Setup}
\label{sec:es}
\noindent \textbf{Dataset.} 
We employ four public datasets to validate the proposed AGGAN.
These datasets contains the faces with different races and they have different illumination, occlusion, pose conditions and backgrounds. 
(i) Large-scale Celeb Faces Attributes (CelebA) dataset \cite{liu2015celeba} has more than 200K celebrity images with complex backgrounds. 
To evaluate the performance of the proposed method under the situation where training data is limited.  
We randomly select 1,000 neutral images and 1,000 smile images as training data, and another 1,000 neutral and 1,000 smile images as testing data. 
(ii) RaFD dataset \cite{langner2010presentation} consists of 4,824 images collected from 67 participants. 
Each participant have eight facial expressions in three  gaze directions, which are captured from three different angles. 
We employed all of the images for diversity expression generation task.
(iii) AR Face \cite{martinez1998ar} contains over 4,000 color images in which only 1,018 images have 4 different facial expressions, \textit{i.e.}, smile, anger, fear and neutral expression.
We employ the images with the expression labels of smile and neutral to evaluate our method.
(iv) Bu3dfe \cite{yin20063d} is a 3D facial expression dataset including 100 subjects with 7 different expression categories. 
We employed the images with smile and neutral expressions as training and testing data.

\noindent \textbf{Parameter Setting.}
For all datasets, images are rescaled to $256 {\times} 256 {\times} 3$ and  we do left-right flip for data augmentation.
For optimization, all baselines and the proposed AGGAN are trained with batch size of 1.
All models were trained for 200 epochs on AR Face, Bu3dfe and RaFD datasets, 80 epochs on CelebA dataset.
For all the experiments, we set $\lambda_{cycle}{=}10$, $\lambda_{gan}{=}0.5$, $\lambda_{pixel}{=}1$ and $\lambda_{tv}{=}1e{-}6$.
We set the number of image buffer to 50 similar in~\cite{zhu2017unpaired}.
We use the Adam optimizer~\cite{kingma2014adam} with the momentum terms $\beta_1$=0.5 and $\beta_2$=0.999.
The initial learning rate for Adam optimizer is 0.0001.
After 100 and 40 epochs for different datasets, the learning rate starts linearly decaying to 0.
For $r$ in Eq.~\eqref{eqn:curriculum}, we set it to 0.01 at the first 10 epochs.
After curriculum period, we set $r$ to 0.5.
The proposed AGGAN is implemented using public PyTorch framework.
To speed up both training and testing processes, we use a Nvidia TITAN Xp GPU. 

\noindent \textbf{Competing Models.} We consider several state-of-the-art cross-domain image generation models as our baselines.
(i) Unpaired image translation method: CycleGAN~\cite{zhu2017unpaired}, DIAT~\cite{li2016deep}, DiscoGAN~\cite{kim2017learning}, DistanceGAN~\cite{benaim2017one}, Dist. + Cycle~\cite{benaim2017one}, Self Dist.~\cite{benaim2017one}, ComboGAN~\cite{anoosheh2017combogan}, StarGAN~\cite{choi2017stargan};
(ii) Paired image translation method: BicycleGAN~\cite{zhu2017toward}, Pix2pix~\cite{isola2016image}, Encoder-Decoder~\cite{isola2016image} and (iii) Label-, mask- or attention-guided image translation method: IcGAN~\cite{perarnau2016invertible}, ContrastGAN~\cite{liang2017generative} and GANimation~\cite{pumarola2018ganimation}. 
Note that the fully supervised Pix2pix, Encoder-Decoder (Enc.-Decoder) and BicycleGAN are trained with paired data on AR Face and Bu3dfe datasets.
Since BicycleGAN can generate several different outputs with one single input image, we randomly select one output from them for fair comparison. 
To re-implement ContrastGAN, we use OpenFace \cite{baltruvsaitis2016openface} to obtain the face masks as extra input data.
For a fair comparison, we implement all the baselines using the same setups as our approach.

\noindent \textbf{Evaluation Metrics.}
We adopt Amazon Mechanical Turk (AMT) perceptual studies to evaluate the generated images.
We gather data from 50 participants per algorithm we tested. 
Participants were shown a sequence of pairs of images, one real image and one fake (generated by our algorithm or a baseline), and asked to click on the image they thought was real.
To seek a quantitative measure that does not require human participation, Mean Squared Error (MSE) and Peak Signal-to-Noise Ratio (PSNR)  are employed.

\subsection{Comparison with the State-of-the-Art}
\noindent \textbf{Bu3dfe\&AR Face Dataset.}  
Results of Bu3dfe and AR Face datasets are shown in Fig.~\ref{fig:bu3dfe_ar}.
It is clear that the results of Dist.+Cycle and Self Dist. cannot even generate human faces.
DiscoGAN produce identical results regardless of the input faces, which suffers from mode collapse.
While the results of DualGAN, DistanceGAN, StarGAN, Pix2pix, Encoder-Decoder and BicycleGAN tend to be blurry.
While ComboGAN and ContrastGAN can produce the same identity but without expression changing.
CycleGAN can generate sharper images, but the details of the generated faces are not convincing.
Compared with all the baselines, the results of our AGGAN are more smooth, correct and with more details.  

\noindent \textbf{CelebA Dataset.}
Since CelebA dataset do not provide paired data, thus we cannot conduct experiments on supervised methods.
The results of CelebA dataset are shown in Fig.~\ref{fig:celeba}. 
We can see that only the proposed AGGAN can produce photo-realistic faces with correct expressions.
Thus, we can conclude that even though the subjects in the three datasets have different races, poses, skin colors, illumination conditions, occlusions and complex backgrounds, our method consistently generates more sharper images with correct expressions than the baseline models. 
Moreover, we observe that our AGGAN preforms better than other baselines when training data is limited, which also shows that our method is very robust.

\begin{table}[!t] \small
	\centering
	\caption{Quantitative comparison with different models. For all metrics except MSE, high is better. }
	\resizebox{1\linewidth}{!}{
		\begin{tabular}{l|c|c|c|c|c|c|c} \toprule
			\multirow{2}{*}{Model}   & \multicolumn{3}{c|}{AR Face}& \multicolumn{3}{c|}{Bu3dfe}  & CelebA \\ \cline{2-8}
			& AMT & PSNR             & MSE              & AMT & PSNR             & MSE            & AMT   \\ \midrule		
			CycleGAN       			  & 10.2  & 14.8142          & 2538.4           & 25.4        & 21.1369          & 602.1          & 34.6     \\ \hline
			DualGAN       			     & 1.3    & 14.7458          & 2545.7           & 4.1          & 21.0617          & 595.1          & 3.2       \\ \hline
			DiscoGAN                 & 0.1    & 13.1547          & 3321.9           & 0.2         & 15.4010          & 2018.7         & 1.2        \\ \hline
			ComboGAN & 1.5     & 14.7465          & 2550.6           & 28.7       & 20.7377          & 664.4          & 9.6       \\ \hline  
			DistanceGAN       & 0.3     & 11.4983          & 4918.5           & 8.9         & 13.9514          & 3426.6         & 1.9     \\ \hline 
			Dist. + Cycle       & 0.1     & 3.8632           & 27516.2          & 0.1           & 10.8042          & 6066.7         & 1.3    \\ \hline 
			Self Dist.         & 0.1     & 3.8674           & 26775.4          & 0.1          & 6.6458           & 14184.2        & 1.2       \\ \hline 
			StarGAN           & 1.6     & 13.5757          & 3360.2           & 5.3           & 20.8275          & 634.4          & 14.8    \\ \hline
			ContrastGAN  & 8.3     & 14.8495          & 2511.1           & 26.2        & 21.1205          & 607.8          & 25.1     \\ \hline
			Pix2pix       & 2.6      & 14.6118          & 2601.3           & 3.8         & 21.2864          & 580.6          & -           \\ \hline
			Enc.-Decoder   & 0.1     & 12.6660          & 3755.4           & 0.2         & 16.5609          & 1576.7         & -        \\ \hline 
			BicycleGAN       & 1.5     & 14.7914          & 2541.8           & 3.2        & 19.1703          & 1045.4          & -         \\ \hline
			\textbf{AGGAN} & \textbf{12.8} & \textbf{14.9187} & \textbf{2508.6}  & \textbf{32.9}  & \textbf{21.3247 }& \textbf{574.5}          & \textbf{38.9}  \\ \bottomrule		
	\end{tabular}}
	\vspace{-0.1cm}
	\label{tab:result}
\end{table}

\begin{table}[!tbp] \small
	\centering
	\caption{Ablation study of AGGAN.} 
	\begin{tabular}{lccc}\toprule
		Component             &  AMT              &   PSNR                  & MSE \\ \midrule
		Full                          &  \textbf{12.8 }&   \textbf{ 14.9187} & \textbf{2508.6}  \\ \hline 
		Full - AD                  &  10.2              &    14.6352              & 2569.2 \\ \hline 
		Full - AD - AG          &  3.2              &   14.4646               & 2636.7 \\ \hline
		Full - AD - PL           &  8.9               &    14.5128              & 2619.8 \\ \hline 
		Full - AD - AL           &  6.3               &   14.6129               & 2652.6 \\ \hline 
		Full - AD - PL - AL    &  5.2               &  14.3287                & 2787.3  \\ \bottomrule  
	\end{tabular}
	\label{tab:abl}
	\vspace{-0.3cm}
\end{table}

\noindent \textbf{RaFD Dataset.} 
Our model can be easily extended to generate facial diversity expressions (\emph{e.g.}, sad, happy and fearful).
To generate diversity expressions in one single model we employ the domain classification loss proposed in StarGAN. 
The results compared against the baselines DIAT, CycleGAN, IcGAN, StarGAN and GANimation are shown in Fig.~\ref{fig:rafd}.
For GANimation, we follow the authors' instruction and use OpenFace \cite{baltruvsaitis2016openface} to obtain the action units of each face as extra input data.
We observe that the proposed AGGAN achieves competitive results compared to GANimation.

\noindent \textbf{Quantitative Comparison on All Datasets.}
We also provide quantitative results on AR Face, Bu3dfe and CelebA datasets.
As shown in Table~\ref{tab:result}, we can see that the proposed AGGAN achieves best results on these datasets compared with competing models including fully-supervised methods Pix2pix, Encoder-Decoder, BicycleGAN, and mask-, label- conditional method, \textit{i.e.}, ContrastGAN.

\subsection{Model Analysis}
\begin{figure}[!t] \tiny
	\centering
	\includegraphics[width=1\linewidth]{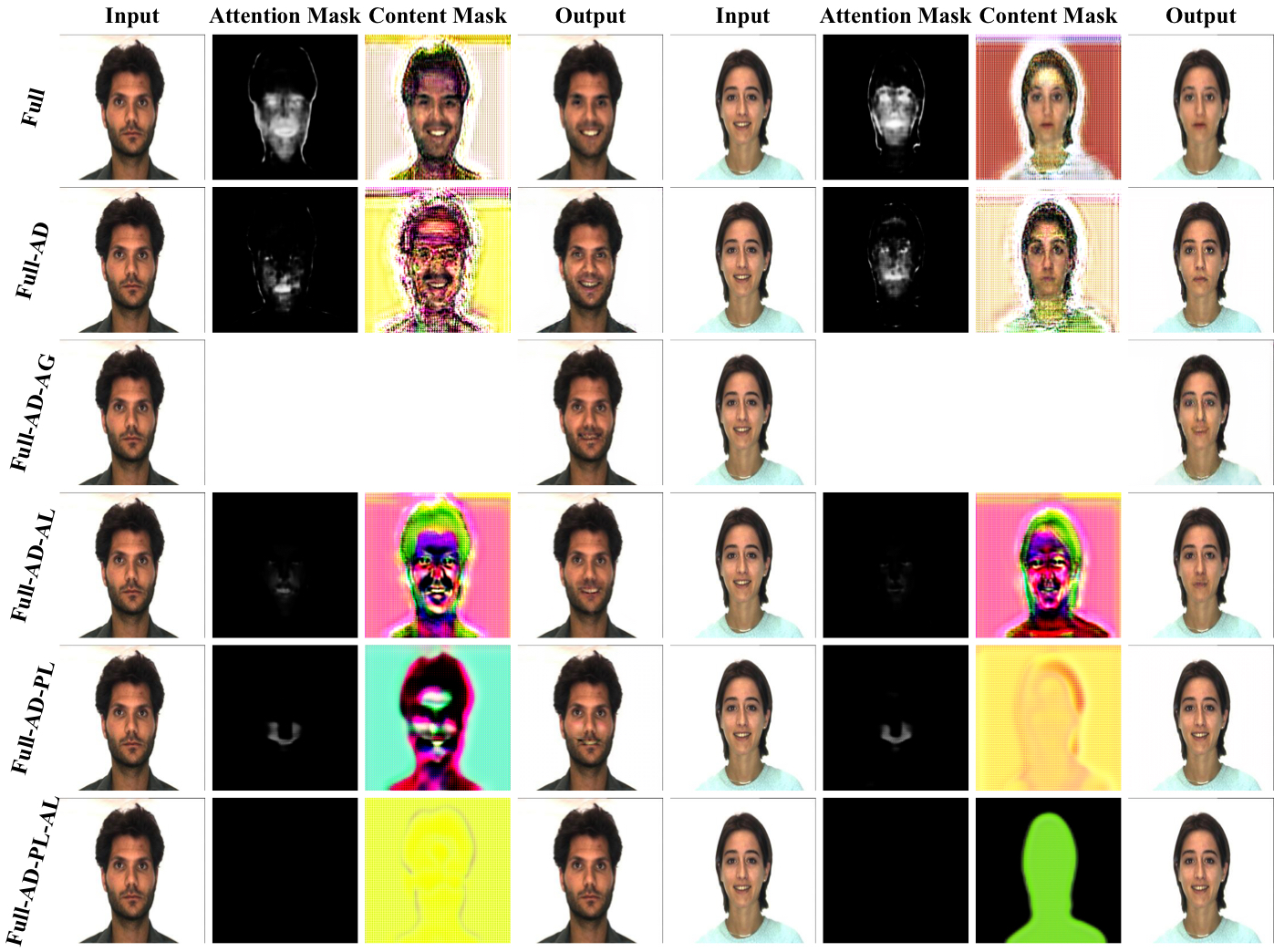}
	\caption{Ablation study on the AR Face dataset.}
	\label{fig:ar_loss} 
	\vspace{-0.4cm}
\end{figure}

\noindent \textbf{Analysis of Model Component.}
In Table \ref{tab:abl} and Fig.~\ref{fig:ar_loss} we run ablation studies of our model on the AR Face dataset.
We gradually remove components of the proposed AGGAN, \emph{i.e.}, Attention-guided Discriminator (AD), Attention-guided Generator (AG), Attention Loss (AL), Pixel Loss (PL).
We find that removing one of them substantially degrades results, which means all of them are critical to our results.
Note that without AG we cannot generate the attention mask and content mask, as shown in Fig.~\ref{fig:ar_loss}

\begin{figure}[!t] \tiny
	\centering
	\includegraphics[width=1\linewidth]{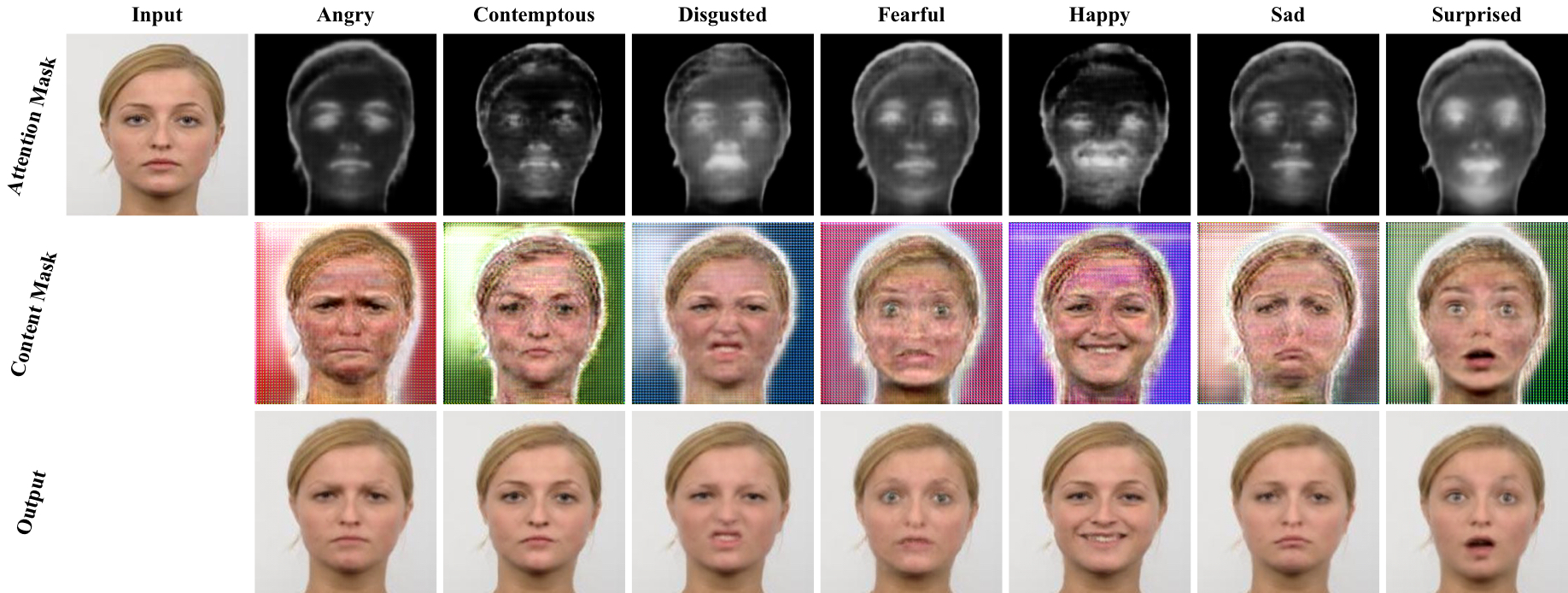}
	\includegraphics[width=1\linewidth]{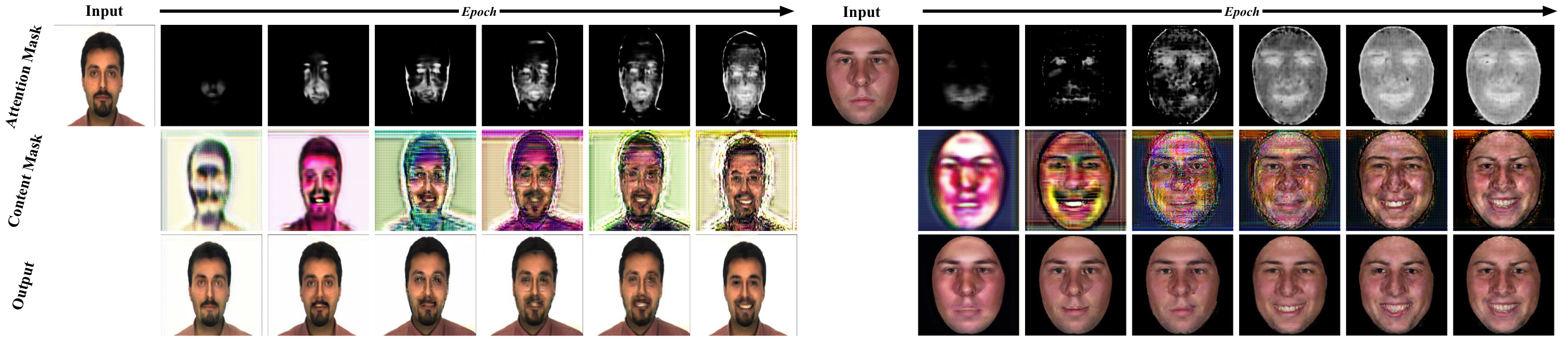}
	\caption{Visualization of attention mask and content attention generation on RaFD (Top), AR Face (Bottom-Left) and Bu3dfe (Bottom-Right) datasets.}
	\label{fig:ar_model} 
	\vspace{-0.4cm}
\end{figure}

\noindent \textbf{Attention/Content Mask Visualization.}
Instead of regressing a full image, our generator outputs two masks, a content mask and an attention mask. 
We also visualize the generation of both masks on RaFD dataset in Fig.~\ref{fig:ar_model}.
We observe that different expressions generate different attention masks and  content masks.
The proposed method makes the generator focus only on those discriminative regions of the image that are responsible of synthesizing the novel expression. 
The attention mask mainly focuses on the eyes and mouth, which means these parts are important for generating the novel expression. 
The proposed method also keeps the other elements of the image  or unwanted part untouched. 
In Fig.~\ref{fig:ar_model} the unwanted part are hair, cheek, clothes and also background, which means these parts have no contribution in generating the novel expression.
Methods without attention cannot learn the most discriminative part and the unwanted part as shown in Fig.~\ref{fig:celeba}.
All existing methods failed to generate the novel expression, which means they treat the whole image as the unwanted part, while the proposed AGGAN can learn the novel expression, by distinguishing the discriminative part from the unwanted part.
Moreover, we also present the  generation of both masks on AR Face and Bu3dfe datasets epoch-by-epoch in Fig.~\ref{fig:ar_model}.
We can see that with the number of training epoch increases, the attention mask and the result become better, and the attention mask correlates well with image quality, which demonstrates that our method is effective.

\begin{table}[!t] \small
	\centering
	\caption{Comparison of the overall model capacity on RaFD Dataset (m=8).}
	\resizebox{0.67\linewidth}{!}{
		\begin{tabular}{l|c|c} \toprule		    
			Method     & \# of Models   & \# of Parameters \\ \midrule
			Pix2pix          & m(m-1)  & 57.2M $\times$ 56 \\
			Encoder-Decoder & m(m-1) &  41.9M  $\times$ 56 \\
			BicycleGAN          & m(m-1)  & 64.3M $\times$ 56 \\   \hline
			CycleGAN & m(m-1)/2        & 52.6M $\times$ 28 \\
			DualGAN  & m(m-1)/2         & 178.7M $\times$ 28 \\
			DiscoGAN & m(m-1)/2        & 16.6M $\times$ 28 \\
			DistanceGAN & m(m-1)/2   & 52.6M $\times$ 28 \\
			Dist. + Cycle & m(m-1)/2   & 52.6M $\times$ 28 \\
			Self Dist. & m(m-1)/2   & 52.6M $\times$ 28 \\ \hline
			ComboGAN & m                 & 14.4M $\times$ 8 \\ \hline
			StarGAN & 1                 & 53.2M $\times$ 1 \\
			ContrastGAN & 1          & 52.6M $\times$ 1 \\ \hline
			AGGAN(Ours)      & 1           & 52.6M $\times$ 1 \\ \bottomrule		
	\end{tabular}}
	\label{tab:computational}
	\vspace{-0.4cm}
\end{table}

\noindent \textbf{Comparison of the Number of Parameters.}
The number of models for different $m$ image domains and the number of model parameters on  RaFD  is shown in Table \ref{tab:computational}.
Note that our performance is much better than these baselines and the number of parameters is comparable with ContrastGAN, while this model requires the object mask as extra data.

\section{Conclusion}
We propose a novel AGGAN for unsupervised image-to-image translation. 
The generators in AGGAN have the built-in attention mechanism, which can detect the most discriminative part of images and produce the attention mask.
Then the attention mask and the input image are combined to generate the targeted images with high-quality.
We also propose a novel  attention-guided discriminator, which only focus on the attended content. 
The proposed AGGAN is trained by an end-to-end fashion.
Experimental results on four  datasets demonstrate that AGGAN and the training strategy can generate compelling results with more convincing details and correct expressions than the state-of-the-art models.

\small\noindent\textbf{Acknowledgements:} We acknowledge the National Institute of Standards and Technology Grant 60NANB17D191 for funding this research. We also acknowledge the gift donation from Cisco, Inc for this research.

{\small
	\bibliographystyle{ieee}
	\bibliography{egbib}
}

\end{document}